\pdfoutput=1
\documentclass[11pt]{article}

\usepackage[final]{acl}

\usepackage{times}
\usepackage{latexsym}

\usepackage[T1]{fontenc}
\usepackage[utf8]{inputenc}

\usepackage{microtype}

\usepackage{inconsolata}

\usepackage{xcolor}
\usepackage{hyperref}
 \definecolor{darkblue}{rgb}{0, 0, 0.5}
  \hypersetup{colorlinks=true, citecolor=darkblue, linkcolor=darkblue, urlcolor=darkblue}

\usepackage{microtype}
\usepackage{arydshln}
\usepackage{booktabs}
\usepackage{graphicx}
\usepackage{subcaption}
\usepackage{xspace}
\usepackage{amsmath}
\usepackage{cleveref}
\usepackage{parskip}
\usepackage{amsmath}
\usepackage{multirow}
\usepackage{threeparttable}
\usepackage{soul}

\usepackage{xstring}
\usepackage{color}

\usepackage{makecell}

\usepackage{annotation}
\usepackage{tikz}
\usepackage{hyperref}

\title{What's under the hood: Investigating Automatic Metrics on Meeting Summarization}

\author{Frederic Kirstein\textsuperscript{1,2}, Jan Philip Wahle\textsuperscript{1}, Terry Ruas\textsuperscript{1}, Bela Gipp\textsuperscript{1} \\
  \textsuperscript{1}University of Göttingen, Germany \\
\textsuperscript{2}\texttt{kirstein@gipplab.org} }
  
\begin{document}
\maketitle
\AddAnnotationRef

\begin{abstract}
Meeting summarization has become a critical task considering the increase in online interactions.
Despite new techniques being proposed regularly, the evaluation of meeting summarization techniques relies on metrics not tailored to capture meeting-specific errors, leading to ineffective assessment.
This paper explores what established automatic metrics capture and the errors they mask by correlating metric scores with human evaluations across a comprehensive error taxonomy.
We start by reviewing the literature on English meeting summarization to identify key challenges, such as speaker dynamics and contextual turn-taking, and error types, including missing information and linguistic inaccuracy, concepts previously loosely defined in the field.
We then examine the relationship between these challenges and errors using human-annotated transcripts and summaries from encoder-decoder-based and decoder-only Transformer models on the QMSum dataset.
Experiments reveal that different model architectures respond variably to the challenges, resulting in distinct links between challenges and errors.
Current established metrics struggle to capture the observable errors, showing weak to moderate correlations, with a third of the correlations indicating error masking.
Only a subset of metrics accurately reacts to specific errors, while most correlations show either unresponsiveness or failure to reflect the error's impact on summary quality.
\end{abstract}

\section{Introduction}
% Motivation
The rise in remote collaboration has increased the need for effective meeting summarization \cite{MrozAV18, PratamaAKD20}, beneficial for participants and non-attendees.
However, established evaluation metrics do not fully capture the challenges of meeting transcripts such as contextual turn-taking and discourse structure \cite{RennardSHV23, KumarK22d}.
The commonly used ROUGE metric \cite{Lin04} has limitations in reflecting summary quality \cite{LiuL08, CohanG16, FabbriKMX21a}, and newer metrics like BERTScore \cite{ZhangKWW20} have not been thoroughly tested for meeting summarization \cite{KirsteinWGR24b}.
Yet, this assessment would be crucial because meeting summarization comes with unique challenges, such as high abstraction, low extraction rate, and complex reasoning \cite{GaoW22}.

% Research Aims
This study examines how automatic metrics relate to human annotations for meeting summarization and what they actually measure in their scores.
We aim to create a unified understanding of the challenges in meeting summarization and the errors that occur when these challenges are unmet.
Using the QMSum dataset \cite{ZhongYYZ21d}, we have experts annotate the challenges in the transcripts and errors in automatically generated summaries using various models, including domain-standard encoder-decoder architectures and notable decoder-only models.
This setup allows us to find connections between challenges, errors, and eight automatic metrics.

The results show problems with current automatic metrics for evaluating meeting summarization. 
\textit{Structural disorganization} errors are often penalized, matching human judgments, but \textit{hallucination} errors are sometimes rewarded.
ROUGE \cite{Lin04} struggles to distinguish the impact of different errors on quality, even though it is good at penalizing omissions.
Surprisingly, about a third of the metrics-error combinations either ignore or reward errors.
For example, Perplexity favors \textit{incorrect references}, and Lens favors \textit{structural disorganization}.
These observations highlight the need for better evaluation methods in meeting summarization.
Our contributions are threefold:
\begin{itemize}
    \item We conduct a comprehensive literature review to identify and unify the challenges specific to English meeting summarization and the types of errors that commonly occur in model-generated summaries.
    \item We build the first direct correlations between the intrinsic challenges and observable errors they induce for encoder-decoder and decoder-only model architectures, providing a framework for tracking their impact.    
    \item We rigorously evaluate the efficacy of nine prevalent automatic metrics, uncovering what they accurately measure, neglect, and their sensitivity to varying error severities, all corroborated by human annotations.  
\end{itemize}

The codebase, annotations, and guidelines are available on GitHub: 

\url{https://github.com/FKIRSTE/emnlp2024-Meeting-Sum-Metrics}.

\section{Methodology}
\label{methodology}
We conducted a focused literature review to identify challenges and observable errors when failing these in English abstractive meeting summarization.
We target papers published between the introduction of the BART model \cite{LewisLGG20a} in 2019, which serves as the typical backbone model for the field, and 2024.
This scope allows us to focus on challenges pertaining to currently used Transformer-based models.
We set additional inclusion criteria to prioritize relevance.
We consider works that introduce novel methodologies or contribute to defining challenges, with a preference for transformer architectures, and exclude only loosely related studies, multi-lingual, multi-modal approaches, non-abstractive, and non-Transformer-based approaches.
We considered publications from 2024, irrespective of their citation index, acknowledging their emergent influence.
The survey follows the PRISMA checklist \cite{PageMBB21} to prevent bias.

Google Scholar is chosen as the primary database for its comprehensive coverage and advanced query capabilities, yielding more works conforming to our review scope than Web of Science and DBLP.

Our search comprises two stages: The initial phase involves broader queries combining "meeting summarization" with adjunctive terms such as \textit{challenge}, \textit{literature review}, or \textit{survey}, while the subsequent refinement uses challenge-related keywords and synonyms.
Papers are ranked based on Google Scholar metrics, considering the top 100 from each query, and serve as primary source for our analysis.

We apply our criteria to the initial pool of 300 papers to select 18 core articles directly addressing challenges and 40 additional papers discussing challenges and errors implicitly through methodological contributions.
Each shortlisted paper undergoes detailed data extraction to identify relevant challenges, methodologies, foundational models, and metrics collated in a structured format for cross-validation.
This approach allows us to construct a comprehensive view of current challenges and observable errors in English meeting summarization.
% The detailed list of selected papers is provided in \Cref{sec:appendix_challenges} and \Cref{sec:appendix_errors}.

\section{Definitions}
\label{lit_rev}
\subsection{Challenges}
\label{sec:challenges}
This section summarizes key challenges in meeting summarization identified in the literature and their implications on the summarization process.

%\subsubsection*{Linguistic Complexity}
% linguistic intricacies
\noindent \textbf{Spoken language.}
Handle colloquialisms, domain-specific terminology, and various forms of linguistic noise, such as false starts, repetitions, and filler words \cite{KoayRDB20d, KumarK22d, AntonyASK23}, which can appear due to the spoken language nature of meetings.
This challenge can affect the accuracy and clarity of the generated summaries.

% linguistic intricacies
\noindent \textbf{Speaker dynamics.}
Accurately distinguish and track different speakers, their utterances, and specific roles (e.g., project manager, applicant), particularly when roles are topic-dependent \cite{KhalifaBM21a, GuTL22, RennardSHV23}.
Failing to do so can introduce biases and result in incomplete or flawed summaries, leaving out noteworthy elements.

% coherence and reference resolution
\noindent \textbf{Coreference.}
Manage the resolution of references to other speakers and previous actions to ensure coherent and complete summarization \cite{LiuSC21}.
Inaccurate handling can lead to ambiguous or incomplete summaries lacking context.

%\subsubsection*{Discourse and Contextual Structure}
% high level structural feature
\noindent \textbf{Discourse structure.}
Understand and track the inherent high-level structure and flow of a meeting \cite{LiHXA23a} throughout the different meeting phases.
These phases may refer to multiple topics \cite{FengFQ22c}, e.g., during an argumentation, or swiftly shift from one topic to another \cite{ShindeGSB22}.
% An agenda may influence the sequencing and prominence of these topics.
Failure to consider this hierarchical structure can result in summaries that are either incomplete or lack coherence.

% local dynamics among individual speakers
\noindent \textbf{Contextual turn-taking.}
Capture the evolving local dynamics of a meeting as it progresses through multiple speaker turns.
The task involves accurately processing shifts in discourse complicated by interruptions, repetitions, and redundancies \cite{MaZEGWS22, ZhangZ21, ShindeGSB22}.
%This is crucial in complex discussions where context and immediate exchanges significantly impact summary quality.
Inadequate capture can result in misleading or shallow summaries.

\noindent \textbf{Implicit context.}
Account for unspoken or implied context, such as tacit organizational knowledge or prior discussions that are not explicitly referred to.
Neglecting this can produce misleading or shallow summaries \cite{XuHPD22}.

% \label{low information density}
\noindent \textbf{Low information density.} 
Identify key points when salient information is sparse and unevenly distributed, especially relevant in decision-making scenarios \cite{RennardSHV23, ZhangNYZ21a}.
This challenge can affect the resulting summary's depth and level of salience.

\noindent \textbf{Data scarcity.}
The lack of diverse, high-quality, real-world scenario training samples hampers model training \cite{RennardSHV23, KumarK22d, JacquenetBL19c}.

\noindent \textbf{Long transcripts.}
Process long transcripts, which can result from longer meetings due to the quadratic computational cost of transformer-based models, leading to efficiency issues with increasing dialogue length \cite{KumarK22d, FengFQ22c, ZhangNYZ21a}.
Only processing a sub-part of a meeting might exclude salient information and, therefore, result in an incomplete or incorrect summary.

\noindent \textbf{Heterogeneous meeting formats.}
Different types of meetings necessitate distinct summarization approaches \cite{RennardSHV23}.
Failure to adapt can result in summaries that miss key points or include irrelevant information. 

\subsection{Error Types}
\label{sec:errors}
Errors in summaries arise when challenges are not correctly addressed.
We have collated error types from the literature into six principal categories, observed in the summaries generated during our experiments as detailed in \Cref{sec:analysis}.

\noindent \textbf{Missing information (MI).}
\noindent This error involves missing information from the meeting, such as significant decisions or actions \cite{ZouSTF23, ChenY20a}.
% while adding irrelevant information
We consider \textit{total omission}, where relevant topics or points discussed are entirely absent from the summary, and \textit{insufficient detail}, where the summary mentions a salient topic but does not capture its depth or the detailed discussion from the original meeting.

\noindent \textbf{Redundancy (Red).}
\noindent The summary contains repeated or redundant information, impacting brevity and clarity \cite{ChenY20a}.
Such repetitions can manifest in different ways: reiterated key points, overuse of individual words, or duplicating entire phrases.

\noindent \textbf{Wrong references (WR).}
\noindent The model misattributes statements, opinions, or actions to incorrect meeting participants or omits their mention altogether \cite{ChenY20a}.

\noindent \textbf{Incorrect reasoning (IR).}
\noindent The model draws conclusions that are not supported by the discussions in the meeting \cite{ChenY20a}.

\noindent \textbf{Hallucination (Hal).}
\noindent The model produces inconsistencies, such as incorrect dates, names, or locations, not aligned with the meeting content \cite{WangZZC22c, JiLFY23}.
This encompasses \textit{extrinsic} hallucinations, introducing events not present in the original meeting or contradicting the input, and \textit{intrinsic} hallucinations, misrepresenting actual events.

\noindent \textbf{Incoherence (Inc).}
\noindent The model generates summaries with disjointed logic or flow.
This manifests as \textit{intra-sentence} disconnections and \textit{inter-sentence} inconsistencies, such as flawed progression and incorrect, contrastive expressions \cite{WangSSJT23}.

We have identified two additional error categories during our human annotation process, as detailed in \Cref{sec:human_annotation}.
Notably, our review of the pertinent literature reveals that these categories have not been explicitly defined in existing studies in a way that matches our findings.

\noindent \textbf{Linguistic inaccuracy (LI).}
\noindent The model uses inappropriate, incorrect, or ambiguous language or fails to capture unique linguistic styles.
This error type spans issues from adopting unsuitable words from the source to grammar mistakes and employing contextually unclear or ambiguous terms.

\noindent \textbf{Structural disorganization (SD).}
\noindent The model may produce summaries that misrepresent the original order or logic of the meeting's discourse, misplacing topics or events.
This error covers only when the order of stated events is wrong.
Including non-existing events or excluding events does not count towards this error.

\section{Experimental Framework}
\begin{table*} [htbp]
  \small
  \centering
    \begin{tabular}{lrrrrr}
    \toprule
    Dataset & \multicolumn{1}{l}{\# Meetings} & \multicolumn{1}{l}{\# Turns} & \multicolumn{1}{l}{\# Speakers} & \multicolumn{1}{l}{avg. Len. of Meet.} & \multicolumn{1}{l}{avg. Len. of Sum.} \\
    \midrule
    AMI  & 137   & 535.6 & 4.0   & 6007.7 & 70.5 \\
    ICSI   & 59    & 819   & 6.3   & 13317.3 & 53.7 \\
    WPCP  & 36    & 207.7 & 34.1  & 13761.9 & 80.5 \\
    \midrule
    all (QMSum) & 232   & 556.8 & 9.2   & 9069.8 & 69.6 \\
    \bottomrule
    \end{tabular}%
  \caption{Statistics for the QMSum subsets and the entire QMSum meeting summarization dataset.}
  \label{tab:statistics_dataset}%
\end{table*}%

\subsection{Models}
\label{sec:models}
We use the Longformer Encoder Decoder (LED) \cite{BeltagyPC20a} as our primary model for its strong summarization performance.
LED employs local+global sliding window attention in the encoder and full self-attention in the decoder, efficiently handling lengthy documents.
To comprehensively assess challenges and errors in encoder-decoder architectures, we further consider DialogLED \cite{ZhongLXZ22c} and PEGASUS-X \cite{PhangZL22a}.
DialogLED extends LED with dialogue-centric pre-training, while PEGASUS-X enhances PEGASUS \cite{ZhangZSL20} for extended inputs using staggered block-local attention.
All models are finetuned on the QMSum training subset (general summaries) for three epochs.

We also explore large language models, using GPT3.5 turbo via ChatGPT and Zephyr-7B-$\alpha$, a refined version of Mistral-7B-v0.1 \footnote{\href{https://huggingface.co/mistralai/Mistral-7B-v0.1}{https://huggingface.co/mistralai/Mistral-7B-v0.1}}.
Mistral's adaptations employ sliding window attention, outperforming Llama2 \cite{TouvronMSA23b}.
To handle context size limitations, we use a chunk-based approach \cite{BhaskarFD23a} with the prompt "Create a TL;DR of the following meeting chunk," inspired by a Microsoft guideline \footnote{\href{https://www.microsoft.com/en-us/microsoft-365-life-hacks/organization/using-chatgpt-creating-meeting-agendas-minutes-notes}{https://www.microsoft.com/en-us/microsoft-365-life-hacks/organization/using-chatgpt-creating-meeting-agendas-minutes-notes}}.
LLMs are used with a zero-shot setup.

For the experiments, the models are grouped into encoder-decoder models containing LED, DialogLED, and Pegasus-X, and the decoder-only models, i.e., GPT-3.5 turbo and Zephyr-7B-$\alpha$.
This grouping is motivated by similar architecture and error distribution and frequency patterns as shown in \Cref{tab:annotation_labels_statistics_errors}.

\subsection{Datasets}
We use as input samples from QMSum \cite{ZhongYYZ21d}, an established dataset for query-based multi-domain meeting summarization.
It includes transcripts from academic (ICSI), product (AMI), and committee meetings (Welsh and Canadian Parliament: WPCP).
ICSI \cite{JaninBEE03} offers informal research meetings with linguistic challenges.
AMI \cite{MccowanCKABF05} provides staged meetings with natural dialogue dynamics.
WPCP presents formal, agenda-driven discussions from UK and Canada Parliament committee meetings.
Detailed statistics are in \Cref{tab:statistics_dataset}.
While there are other datasets out there in the context of meeting summarization, such as the frequently used MeetingBank \cite{HuGDD23b} and ELITR \cite{NedoluzhkoSHG22a} datasets, we do not include these.
MeetingBank is conceptually close to the WPCP meetings, therefore not enhancing the diversity of our data selection and potentially introducing an imbalance in meeting type.
ELITR aims to produce meeting minutes with bullet points containing key insights and actions discussed.
Therefore, it does not fit the abstractive summaries we aim to investigate.

\begin{table*}[]
\small
\centering
\begin{tabular}{l c c c c c c c c c }
\toprule
 & \makecell{Spoken \\ Language} & \makecell{Speaker \\ Dynamic} & \makecell{Co- \\reference} & \makecell{Discourse \\ Structure} & \makecell{Contextual \\ Turn-Taking} & \makecell{Implicit \\ Context} & \makecell{Data \\ Scarcity} & \makecell{Low Inf. \\ Density} & \makecell{Overall} \\
\midrule
detection & 0.84 & 0.77 & 0.68 & 0.79 & 0.73 & 0.90 & 0.87 & 0.78 & 0.79 \\
frequency & 0.84 & 0.71 & 0.78 & 0.70 & 0.89 & 0.91 & 0.90 & 0.84 & 0.82 \\
\bottomrule
\end{tabular}
\caption{Krippendorff's alpha for inter-annotator agreement on challenges and their frequencies, with challenges (abbreviations) in top row ordered as in \Cref{sec:challenges}.}
\label{tab:krippendorffs_alpha_challenge}
\end{table*}

\begin{table*}[]
\small
\centering
\begin{tabular}{l l c c c c c c c c c }
\toprule
&  & MI & Red & WR & IR & Hal & Inc & LI & SD & overall\\
\midrule
detection & Encoder-Decorder & 0.74 & 0.875 & 0.83 & 0.87 & 0.79 & 0.61 & 0.86 & 0.71 & 0.78 \\
Decoder-only & 0.76 & 0.87 & 0.85 & 0.79 & 0.77 & 0.90 & 0.90 & 0.71 & 0.82 \\
\midrule
error impact & (LED) & 0.82 & 0.92 & 0.85 & 0.73 & 0.80 & 0.89 & 0.87 & 0.79 & 0.83 \\
\bottomrule
\end{tabular}
\caption{Krippendorff's alpha for inter-annotator agreement on errors for encoder-decoder (i.e., LED, DialogLED) and decoder-only models (i.e., GPT-3.5, Zephyr-7B-$\alpha$). For the LED model, the agreement on error impact is also reported. Errors (abbreviations) in the top row are ordered as in \Cref{sec:errors}.}
\label{tab:krippendorffs_alpha_error}
\end{table*}

\subsection{Metrics}
We align human annotations with automatic metrics considering count-based, model-based, and QA-based methodologies, chosen based on their prevalent use in meeting and dialogue evaluation \cite{GaoW22}.

\paragraph{Count-based.}
ROUGE \cite{Lin04} is the default-used metric suite that assesses the overlap between n-grams in generated summaries that appear in the reference.
Researchers mainly consider unigrams, bigrams, and the longest common sequence to gauge the relevance and accuracy of generated content.
BLEU \cite{PapineniRWZ02} evaluates how many n-grams from the reference appear in the generated summary.
The score is designed to reflect the precision of the generated text, focusing primarily on lexical similarity to the reference.
METEOR \cite{BanerjeeL05} builds on BLEU by accounting for synonyms, word stems, and sentence structure, offering a holistic assessment of lexical, syntactic, and semantic alignment in precision and recall.

\paragraph{Model-based.}
BERTScore \cite{ZhangKWW20} measures the contextual similarity between generated and reference texts using a pre-trained BERT model, reflecting semantic and syntactic similarity \footnote{We report the rescaled BERTScore-F score: \newline
\href{https://github.com/Tiiiger/bert_score/blob/master/journal/rescale_baseline.md}{https://github.com/Tiiiger/bert score/blob/master/journal/rescale baseline.md}}.
Perplexity (PPL) measures a language model's uncertainty in predicting words, evaluating the quality and fluency of utterances.
We use GPT-2 \cite{RadfordWCL19} as the language model.
BLANC \cite{LitaRL05} measures how well a generated summary aids a language model in understanding the original document, reflecting informativeness.
LENS \cite{MaddelaDHX23a} is a trainable metric that assesses the alignment of generated text with human references in content and style. While this metric is also explored for meeting summarization, it is noteworthy that it is based on RoBERTa, adopting its maximum context length of 512 tokens, making it technically less suitable for assessing meeting summaries considering transcripts.

\paragraph{QA-based.}
QuestEval \cite{ScialomDLP21} combines FEQA \cite{DurmusHD20} and SummaQA \cite{ScialomLPS19} scores, using a question-answering model to answer questions formed from the reference text (SummaQA) or the generated summary (FEQA), extracting answers from the opposite source.
QuestEval evaluates factuality, coherence, informativeness, and relevance.

\paragraph{Human annotation.}
\label{sec:human_annotation}
We adapt proven methodologies \cite{ZhangLYF23} for a thorough annotation process involving four graduate students from diverse backgrounds (i.e., computer science, psychology, communication science), all well-versed in English and familiar with meeting summarization.
From the QMSum general-summary test set, we choose 35 general-summary samples, each containing meeting transcripts, gold summaries, and model-generated summaries, resulting in 175 distinct samples for annotation.
Annotators identify challenges (\Cref{sec:challenges}) in the transcript and errors \Cref{sec:errors} in the generated summaries using yes/no questions such as: "Does the given summary omit crucial information or provide insufficient detail about salient points?" corresponding to the \textit{missing information} error.
The annotators further rate the frequency of challenges and the impact of errors for the LED model using Likert scales from rare occurrence or minimal impact (1) to frequent presence or high impact (5).

To ensure reliability and consistency, we assess inter-annotator agreement using Krippendorff's alpha, achieving an average of 0.81 (see \Cref{tab:krippendorffs_alpha_challenge,tab:krippendorffs_alpha_error}).
A preliminary pilot test serves as annotator training and refinement of guidelines.
Regular review meetings are held to maintain consistency.
Annotators also highlight summary segments with errors.
Discrepancies are discussed, and an expert annotator is available to discuss complex issues.

A full presentation of the annotation process is stated in \Cref{sec:appendix_annotation_process}.
Details on annotated labels are shown in \Cref{sec:appendix_annotation_labels}.

\section{Analysis}
\label{observing_challenges}
\label{sec:analysis}

\subsection{Linking challenges and error types}
\begin{table*}[h]
\centering
\scriptsize
\begin{tabular}{lp{55mm}p{55mm}}
\toprule
\textbf{Errors} & \textbf{Encoder-Decoder} & \textbf{Decoder-only}\\
\midrule
missing information & discourse structure*, implicit context & coreference*\\
\midrule
redundancy & speaker dynamics, contextual turn-taking* \newline implicit context*, decision dynamics* & spoken language*, speaker dynamics \newline contextual turn-taking, low info density\\
\midrule
wrong references & (none) & spoken language**, contextual turn-taking* \newline low information density**\\
\midrule
incorrect reasoning & speaker dynamics, implicit context \newline low information density & (none)\\
\midrule
hallucination & contextual items, implicit context* & coreference*, contextual turns* \newline implicit context\\
\midrule
incoherence & spoken language*, coreference* \newline contextual turn-takin*, decision dynamics* & coreference\\
\midrule
linguistic inaccuracy & coreference, low information density & spoken language*, speaker dynamics** \newline decision dynamics*, low information density \\
\midrule
structural disorganization & spoken language**, speaker dynamics** \newline contextual turns*, decision dynamics \newline low information density & coreference*, contextual turns \newline implicit context*, decision dynamics\\
\bottomrule
\end{tabular}
\caption{
% based on the correlation matrix in \Cref{sec:appendix_corr_sig_matrices}.
The linkage between challenges and errors. No asterisk indicates low correlation, * signifies mid correlation (p $\leq$ 0.05), and ** denotes high correlation (p $\leq$ 0.01).}
\label{tab:link_challenge_error}
\end{table*}

We analyze the relationship between challenges and frequently observed errors for encoder-decoder and decoder-only architectures using Point-biserial correlation based on human annotations from \Cref{sec:human_annotation}. 
\Cref{tab:link_challenge_error} presents the results.

Encoder-decoder models exhibit strong links between \textit{incoherence}, \textit{structural disorganization}, and \textit{redundancy} errors and challenges like \textit{spoken language} and \textit{speaker dynamics}, suggesting struggles with maintaining coherence.
Conversely, \textit{wrong references}, \textit{linguistic inaccuracy}, and \textit{hallucination} errors show limited associations with the examined challenges.
Specifically, the limited association of \textit{hallucination} reinforces current understandings, suggesting that the underlying causes remain elusive \cite{MaynezNBM20a}.
Notably, the \textit{implicit context} challenge works as a proxy for \textit{hallucination}.
The minimal correlation of the \textit{wrong reference} error, which rarely occurs due to the sentence structure in generated summaries, and the \textit{linguistic inaccuracy} error indicate that challenges such as \textit{coreference} are well-handled by these models.

Regarding specific challenges, the \textit{discourse structure} challenge correlates slightly with the \textit{missing information} error, indicating occasionally missed details within distinct phases.
The \textit{low information density} challenge weakly correlates with the \textit{incorrect reasoning} error, revealing difficulties in extracting salient details.
The \textit{contextual turn-taking} challenge aligns with \textit{redundancy} and \textit{incoherence} errors, suggesting issues in capturing dynamics on a granular level within the different meeting phases.

Decoder-only models show different patterns, with \textit{incorrect reasoning} being the rarest and weakest correlated error.
\textit{Wrong references}, \textit{redundancy}, and \textit{linguistic inaccuracy} errors are most prevalent, aligning with \textit{low information} and \textit{spoken language} challenges.
For LLMs, redundancies manifest as repetitive introductory sentences, while linguistic inaccuracies emerge as contextually ambiguous terms.
\textit{Missing information} and \textit{structural disorganization} errors are equally prominent and strongly correlated, linking to the \textit{coreference} challenge and emphasizing models' tendencies to list topics without proper context.
Some negative correlations suggest specific challenges might decrease the likelihood of certain errors, such as for \textit{low information density} challenge and \textit{missing information} error, warranting further exploration.

% RQ1: Do specific errors tend to occur only when challenges reach a certain frequency level?
% > Conditional frequency analysis

% RQ2: What is the impact level of these errors on the frequency of the challenges?

\subsection{Correlation of automatic metrics and human annotation}
Though numerous and frequently applied, existing automatic metrics are not tailored to the intricacies of meeting summarization.
We analyze their reactions to the diverse error types observable in meeting summaries by answering the following three research questions.
% A comparison between human annotations and automatic metrics is detailed in \Cref{tab:human_annotation} under \Cref{sec:appendix_manual_inspection}.

\paragraph{RQ1: How do automatic metrics correlate with human assessments?}
\begin{table*}[]
\small
\centering
\begin{tabular}{l c c c c c c c c c c}
\toprule
 & MI & Red & WR & IR & Hal & Inc & LI & SD \\
\midrule
ROUGE-1 & \textbf{-0.40*} & 0.17 & -0.07 & -0.18 & 0.05 & \textbf{-0.30} & \textbf{-0.12} & \textbf{-0.41*} \\

ROUGE-2 & -0.20 & 0.10 & 0.07 & \textbf{-0.29} & \textbf{0.10} & \textbf{-0.22} & \textbf{-0.10} & -0.35* \\

ROUGE-L & \textbf{-0.29} & 0.08 & -0.04 & -0.21 & 0.02 & -0.21 & -0.09 & \textbf{-0.46**} \\

% rougeLsum & \textbf{-0.27} & -0.30 & 0.04 & 0.19 & -0.10 & -0.14 & 0.12 & -0.18 & -0.18 & -0.32 \\

BLEU & -0.20 & 0.08 & -0.09 & -0.08 & \textbf{0.32} & -0.13 & 0.05 & -0.26 \\

METEOR & \textbf{-0.26} & \textbf{0.27} & \textbf{0.16} & \textbf{-0.23} & 0.08 & -0.20 & 0.02 & -0.38* \\

% CHRF & \textbf{-0.29} & -0.25 & -0.08 & \textbf{0.31} & 0.14 & -0.15 & 0.23 & -0.12 & -0.11 & -0.30 \\

% BERTScore precision & -0.17 & \textbf{-0.41*} & \textbf{0.24} & 0.17 & 0.05 & 0.02 & 0.09 & -0.18 & -0.10 & -0.27 \\

% BERTScore  recall & -0.13 & -0.24 & -0.08 & \textbf{0.25} & 0.11 & -0.13 & 0.09 & 0.15 & 0.08 & -0.21 \\

\midrule

BERTScore (F) & -0.16 & \textbf{0.22} & 0.09 & -0.06 & \textbf{0.10} & -0.02 & -0.01 & -0.26 \\

PPL & -0.10 & -0.10 & \textbf{0.44**} & \textbf{-0.32} & -0.08 & 0.09 & \textbf{0.24} & -0.17 \\

BLANC (TS) & -0.19 & 0.05 & -0.13 & -0.13 & -0.13 & \textbf{-0.42*} & -0.09 & -0.36* \\

LENS & 0.01 & \textbf{-0.38*}& \textbf{-0.17} & 0.20 & 0.01 & 0.03 & -0.03 & \textbf{0.45**} \\

\midrule

QuestEval (F) & 0.10 & -0.05 & \textbf{0.16} & 0.02 & 0.26 & 0.00 & -0.03 & -0.28 \\

% QuestEval Precision & 0.07 & -0.18 & \textbf{0.26} & 0.01 & \textbf{0.18} & 0.07 & 0.27 & -0.08 & -0.09 & -0.31 \\

% QuestEval Recall & 0.13 & -0.02 & \textbf{-0.15} & -0.17 & 0.03 & -0.11 & 0.11 & 0.19 & 0.13 & -0.06 \\
\bottomrule
\end{tabular}
\caption{Point-biserial correlation between automatic metrics and annotated errors on samples generated by LED. * denotes significance at p $\leq$ 0.05 and ** at p $\leq$ 0.01. The top row shows abbreviated error types from \Cref{sec:errors} in order. A negative correlation indicates worsening metric scores with increasing error occurrence. The three highest absolute values are highlighted in \textbf{bold}. We present F-scores (F) for BERTScore and QuestEval, with BERTScores being further rescaled.
}
\label{tab:correlation_metric_error}
\end{table*}

\Cref{tab:correlation_metric_error} shows Point-biserial correlations between automatic metrics and expert-annotated errors for LED model-generated summaries.
Our analysis reveals that no metric consistently correlates highly with all error types, underscoring the complexity of meeting summarization and the absence of a universal metric that captures human judgment well \cite{GaoW22}.
However, some metrics show trends aligning with human judgment through significant negative scores.
Though not designed for structural coherence, several metrics show significant negative correlations with \textit{structural disorganization} errors, indicating that temporal and logical disorganization breaks n-gram sequences and semantic flow.
ROUGE-1 exhibits a more significant score than ROUGE-2 and ROUGE-L, aligning with observations in dialogue summarization \cite{GaoW22}.
In particular, the gaps, disjoint narratives, and incorrect statements from errors such as \textit{missing information}, \textit{incoherence}, and \textit{incorrect reasoning} influence metric scores by not aligning with reference n-grams or shifting meaning \cite{Lin04}. 
\textit{Redundancy}, \textit{wrong references}, and \textit{hallucination} errors remain less frequently detected, as summaries with these errors can still closely match the reference in terms of n-gram overlap and semantic similarity.

As expected, count-based metrics are responsive to \textit{missing information} \cite{Lin04}, with ROUGE-1 correlating significantly with information omissions \cite{GaoW22} and BLEU identifying summaries with key content omissions.
Count-based metrics significantly correlate with \textit{structural disorganization} errors though not designed to detect this error, possibly because of disrupted n-gram sequences due to reorganized content \cite{BanerjeeL05}. 
ROUGE and Meteor are potential indicators of incorrect reasoning and incoherence.

Among model-based metrics, BERTScore exhibits sensitivity to \textit{missing information}, reflecting semantic and contextual differences between candidate and gold summaries \cite{ZhangKWW20}.
However, its correlation with \textit{structural disorganization} is less direct, hinting at the influence of sentence alignment and structural coherence.
BLANC, designed for evaluating coherence and fluency, correlates as expected negatively with incoherence.
LENS effectively captures redundancy errors.
Model-based metrics present milder correlations, possibly due to discrepancies between training data and meeting contexts \cite{GaoW22}.
While these metrics seem to not align as well with human judgment as count-based metrics, they offer insights into errors not captured by count-based metrics.

The correlation table shows that no single metric predominantly captures all error types, with most correlations being weak to moderate. 
Examining \Cref{tab:correlation_metric_error}, a combination of metrics like ROUGE, BLANC, and LENS could serve as a proxy for various errors, but this approach warrants additional evaluation and score weighting.

\begin{table*}[]
\small
\centering
\begin{tabular}{l c c c c c c c c c c}
\toprule
 & MI & Red & WR & IR & Hal & Inc & LI & SD \\
\midrule
ROUGE-1 & 0.07 & \textbf{0.24} & -0.02 & -0.17 & 0.10 & \textbf{-0.30} & -0.09 & \textbf{-0.42*} \\

ROUGE-2 & -0.05 & 0.15 & 0.06 & \textbf{-0.29} & 0.13 & \textbf{-0.23} & \textbf{-0.11} & -0.36* \\

ROUGE-L & 0.03 & 0.15 & -0.05 & -0.22 & 0.10 & -0.21 & -0.08 & \textbf{-0.44**} \\

BLEU & -0.11 & 0.08 & -0.06 & -0.09 & \textbf{0.35*} & -0.14 & 0.03 & -0.31 \\

METEOR & -0.06 & \textbf{0.21} & 0.16 & \textbf{-0.27} & 0.13 & -0.21 & -0.05 & -0.36* \\

\midrule

BERTScore (F) & \textbf{-0.25} & 0.19 & 0.08 & -0.10 & \textbf{0.15} & -0.02 & -0.04 & -0.31 \\

PPL & -0.02 & -0.13 & \textbf{0.42} & \textbf{-0.29} & 0.02 & 0.00 & \textbf{0.23} & -0.13 \\

BLANC & -0.03 & 0.04 & -0.07 & -0.15 & -0.11 & \textbf{-0.44} & \textbf{-0.13} & -0.36* \\

LENS & \textbf{-0.14} & \textbf{-0.42*} & \textbf{-0.26} & 0.22 & -0.07 & 0.00 & -0.06 & \textbf{0.40} \\

\midrule

QuestEval (F) & \textbf{0.13} & 0.04 & \textbf{0.18} & 0.05 & \textbf{0.34*} & -0.03 & -0.07 & -0.24 \\
\bottomrule
\end{tabular}
\caption{Spearman correlation between metrics and annotated error impacts, using summaries generated by LED as base. * denotes significance at p $\leq$ 0.05 and ** at p $\leq$ 0.01. The top row lists abbreviated error types from \Cref{sec:errors} in sequence. A negative correlation implies declining metric scores with rising error instances. The three most pronounced absolute values are emphasized in \textbf{bold}. F-scores (F) are given for BERTScore and QuestEval, with BERTScores being further rescaled.
}
\label{tab:correlation_metric_error_impact}
\end{table*}

\paragraph{RQ2: Do automatic metrics mask errors?}

We categorize "masking" as indifference to an error (near-zero correlation) or positive reinforcement of a mistake (positive correlation).
Count-based metrics predominantly show near-zero or negative correlations with errors, indicating they might not sufficiently penalize specific mistakes. 
This behavior is expected \cite{SaadanyO21,AkterBK22}, especially for errors these metrics were not explicitly designed to detect or for which they can only act as a proxy.

From \Cref{tab:correlation_metric_error} using summaries generated by LED as a base, we observe that some metrics show their known struggle with specific error types through near-zero scores (e.g., BERTScore and linguistic inaccuracy \cite{HannaB21}), model-based metrics occasionally manifest positive correlations.
%QuestEval rewards errors like \textit{missing information} and \textit{hallucination}, a counterintuitive outcome given its focus on factual accuracy.
%The finding indicates that hallucinated details likely bypass metric detection due to generated questions lacking the rigor to spot them, which can be linked to the gap between pre-training and meeting data \cite{GaoW22}.
%BERTScore, emphasizing semantic and syntactic similarities, may reward redundant yet correct details \cite{HannaB21}.
Perplexity favors incorrect references if they preserve linguistic coherence and fluency, aligning with the language model's expectations and yielding a lower (better) score.
LENS appears to struggle with capturing broader logical and temporal structures.
Despite being designed for text simplification, it significantly correlates with the \textit{structural disorganization} error and primarily emphasizes word-level and intra-sentence simplification \cite{MaddelaDHX23a}.
While QuestEval and BETScore do not exhibit significant correlations, they show noteworthy trends:
QuestEval may reward errors like \textit{missing information} and \textit{hallucination}, a counterintuitive outcome given its focus on factual accuracy.
The finding could indicate that hallucinated details likely bypass metric detection due to generated questions lacking the rigor to spot them, which can be linked to the gap between pre-training and meeting data \cite{GaoW22}.
BERTScore, emphasizing semantic and syntactic similarities, may reward redundant yet correct details \cite{HannaB21}.

\paragraph{RQ3: How do the metric scores vary with the severity of the error?}

To identify the relationship between metric scores and error severity, we analyze how these scores fluctuate by the impact of errors.
\Cref{tab:correlation_metric_error_impact} shows the Spearman correlation trends, using summaries generated by LED as a base.
Across different models and error severities, most correlations are negligible to weak, and many lack statistical significance.
This observation emphasizes the limitations of current metrics in discerning error severity in meeting summarization.

BLEU and QuestEval display significant positive correlations with \textit{hallucination} errors, underscoring their vulnerability to hallucinated content.
While metrics such as BERTScore and LENS show sensitivity to \textit{missing information} in RQ1, their precision in assessing error severity is limited.
This limitation may stem from the gap between training data and meeting transcripts for the metrics utilizing language models, as the meeting transcript was provided as part of the input \cite{GaoW22}.
LENS, closely aligned with text simplification, penalizes \textit{redundancy}. %, while Perplexity is particularly sensitive to \textit{wrong references} and \textit{linguistic inaccuracies}.
These observations underscore that some metrics are intrinsically responsive to specific errors, even if not explicitly designed for them.
Perplexity does not exhibit significant correlations.

\section{Related Work}
Research on meeting summarization typically addresses challenges and error types in isolation, leaving their interrelation unexplored.
For dialogue summarization, challenges are collected \cite{ChenY20a} but lack the detail required for meeting contexts.
Previous works by \citet{TangNWW22b, WangZZC22c} lay the groundwork for our error typologies, enhanced with insights from diverse studies and annotator feedback.
\citet{ChenY20a} correlate challenges and errors in dialogue summarization, but their findings do not entirely transfer to English abstractive meeting summarization, though their annotation approach informs ours.
Automatic metrics for meeting summaries are underexplored, but similar studies exist for text summarization using the CNN/Dailymail dataset \cite{FabbriKMX21a} and dialogue summarization via the SAMSum dataset \cite{GaoW22}.
We build upon these works by selectively adopting their metrics and augmenting them with measures commonly reported in recent meetings and dialogue summarization studies.

\section{Conclusion}
We developed a resource suite for meeting summarization evaluation, covering domain-specific challenges, linked error types, predictive correlations for model architectures with known challenges, and expert annotations of the QMSum subset.
Our analysis highlighted the limitations and misalignments of current metrics with human judgment in discerning error nuances. 
Metrics that work well for other summarization tasks either did not react to errors or cannot reflect the impact on quality in their scores.
A composite metric may be more effective, but its formulation requires further research.
Recent advancements highlighted the potential of LLM-based metrics.
Utilizing an LLM prompted with our detailed annotator guidelines and supplemented by examples presents a promising approach for detecting complex errors like structural disorganization and incorrect reasoning.
% Future directions include augmenting the entire QMSum dataset with our annotations, drawing inspiration from the Stanford Context Word Similarity dataset's impact on word sense ambiguity \cite{HuangSMN12}.
% An enriched QMSum may offer deeper insights into potential challenges for new techniques.
% To study our findings across languages, it is of interest to investigate additional datasets (e.g., MeetingBank) and multilingual datasets (e.g., ELITR, FREDSum).
Given the current evolution of techniques, we plan to extend the work as a (dialogue) summarization benchmarking pipeline to capture better how upcoming techniques handle challenges and how well new metrics capture their performance.
We encourage the research community to contribute model outputs and introduce new metrics to this initiative.

\section*{Acknowledgements}
This work was supported by the Lower Saxony Ministry of Science and Culture and the VW Foundation.
Frederic Kirstein was supported by the Mercedes-Benz AG Research and Development.

\section*{Limitations}
Our study, while offering a comprehensive analysis of challenges and errors in meeting summarization, primarily focuses on English-language summaries.
This linguistic focus may lead to variations in challenges and errors across languages.
Some challenges, especially those subtle to human perception, might be overlooked, creating potential gaps in our annotations.
Our reliance on Google Scholar, despite its broad coverage, has its criticisms, as it tends to favor citation counts \cite{Fagan_2017} and may include less reputable sources \cite{Beall98}.

The QMSum dataset, with its 35 samples, provides statistical significance but represents only a fraction of potential meeting types.
Thus, findings may vary across different meeting contexts.

We include a total of 175 samples, comparable to the original QMSum dataset (i.e., 232 samples), one of the most used datasets in meeting summarization.
The considered dataset maintains a diverse representation of meeting types (academic, business, parliament) within the dataset.
We acknowledge the limitation that QMSum might not cover all meeting types in existence. 
However, we focus on creating a unified understanding of the challenges in meeting summarization and the errors that occur when these challenges are unmet.
For that, we used QMSum as a proxy.

The challenges we associate with QMSum are inferred from human annotations rather than directly tested, and our model selection, while reflective of the current landscape, does not capture every variant.
Our choice of encoder-decoder models is comprehensive, but we miss out on architectures like hierarchical models due to accessibility issues, which hamper the comparability of the models.
The array of large language models is continually expanding, and our selection is based on the standings on the Huggingface LLM Leaderboard at the time of writing.
Our method of linking challenges to errors is holistic, yet it might dilute strong connections between them, as we did not test the linking with isolated challenges.
Specific metrics, like LENS and QuestEval, may have biased scores since they use the meeting transcript as input and are not domain-trained.
Lastly, our findings, rooted in the nuances of meeting summarization, might not seamlessly apply to broader summarization domains like dialogue summarization, given each domain's distinctive traits.

\section*{Ethics Statement and Broader Impact}
Our research abides by ethical guidelines for AI research and is committed to privacy, confidentiality, and intellectual property rights.
We have ensured that the datasets in our study, which are publicly available, do not house sensitive or personal details.
While our study leverages existing resources and generative models, it is important to note that these models can possess biases and may occasionally generate summaries with distortions, biases, or inappropriate content \cite{Gooding22}. We have configured our models to omit potentially harmful or unsafe content to counteract this.
While our research aims to enhance meeting summarization to benefit communication and productivity across sectors, we are acutely aware of the ethical challenges posed by AI in this domain.
Meeting summarization models must be wielded with respect to privacy and consent, especially when processing sensitive or confidential material.
It's paramount that these models neither violate privacy nor perpetuate harmful biases.
As the field evolves, we stress the importance of maintaining these ethical considerations and encourage fellow researchers to uphold them, ensuring that AI advancements in meeting summarization are both beneficial and ethically grounded.
An integral aspect of our ethical commitment is reflected in our approach to annotator recruitment and management.
The team of annotators, consisting of interns, student assistants, and doctoral students, was meticulously selected through internal channels.
This strategy was chosen to uphold a high standard of annotation quality—a quality we found challenging to guarantee through external platforms such as Amazon Mechanical Turk.
Ensuring fair compensation, these annotators were reimbursed following institutional guidelines for their respective positions.
Further, flexibility in the annotation process was also a priority.
Annotators were free to choose their working times and environments to prevent fatigue from affecting their judgment.

% \nocite{*}
%\section{Bibliographical References}\label{sec:reference}

\bibliography{custom, references_zotero, zotero, 24_EMNLP_Metrics, addendum}
%\bibliographystyle{lrec-coling2024-natbib}
% \bibliography{lrec-coling2024-example}

\appendix
\section{Annotation process details}
\label{sec:appendix_annotation_process}

Following, we describe the details of our annotation process.

\paragraph{Annotator selection:} 
Our annotation team comprises four graduate students, officially employed as interns or doctoral candidates through standardized contracts.
We select them from a pool of volunteers based on their availability to complete the task without time pressure and their English proficiency (native speakers or C1-C2 certified).
This ensures they can comprehend meeting transcripts, human-written gold summaries from QMSum, and model-generated summaries. We aimed for gender balance (two male, two female) and diverse backgrounds, resulting in a team of two computer science students, one psychology student, and one communication science student, aged 24-28.

\paragraph{Preparation:}
We prepare a comprehensive handbook for our annotators, detailing the project context and defining challenges and error types.
Each definition includes two examples: one with minimal impact (e.g., slight information redundancy) and one with high impact (e.g., repeated information throughout).
The handbook explains the two-part rating system: a binary yes/no for the existence of a challenge or error, followed by a 1-5 Likert scale impact/frequency rating if a characteristic is observed as existing.
This impact/frequency scoring is only used for the challenges (frequency, 1 low - 5 high) and errors (impact, 1 low - 5 high) produced by the LED model.
Annotators are further tasked to provide reasoning for each decision.
The handbook does not specify an order for processing errors or challenges.
We provide the handbook in English and the annotators' native languages, using professional translations. 
The handbook could be used throughout the whole annotation process as a reference.

We set up a five-week timeline for the annotation process, preceded by a one-week onboarding period.
The first two weeks feature twice-weekly check-ins with annotators, which are reduced to weekly meetings for the following three weeks. 
Separate quality checks without the annotators are scheduled weekly.
(Note: week refers to a regular working week)

\paragraph{Onboarding:}
The onboarding week is dedicated to getting to know the project and familiarizing with the definitions and data.
We begin with a kick-off meeting to introduce the project and explain the handbook, mainly focusing on each definition.
We note initial questions to revise the handbook potentially.
Annotators are provided with 35 samples generated by SLED+BART \cite{IvgiSB22}, chosen for their balance of identifiable errors and good-quality summaries while capable of processing the whole meeting.
After the first 15 samples, we hold individual meetings to clarify any confusion and update the guidelines accordingly.
The remaining 20 samples are then annotated using these updated guidelines. 
A second group meeting this week addresses any new definitional issues.
After the group meeting, we meet individually with annotators to review their work, ensuring quality and understanding of the task and samples.
All four annotators demonstrate reliable performance and good comprehension of the task and definitions, judging from the reasoning they provided for each decision and annotation. 
We computed an inter-annotator agreement score using Krippendorff's alpha, achieving 0.81, indicating sufficiently high overlap.

\paragraph{Annotation Process:}
We continue the annotation process similarly.
Each week, we distribute all 35 samples generated by one model to one of the annotators. 
Consequently, one annotator works through all samples of one model in one week.
After five weeks, all samples have been processed by all annotators.
Annotators are unaware of the summary-generating model and are given a week to complete their set at their own pace and break times.
Quiet working rooms were provided if needed for concentration. 
To mitigate position bias, the sample order is randomized for each annotator.
Annotators can choose their annotation order for each sample and are allowed to revisit previous samples to adapt ratings.
To simplify the process, we frame each error type as a question, such as "Does the summary omit crucial information or provide insufficient detail about salient points?" for the missing information error.

Regular meetings are held to address questions on definitions or emerging issues.
During the quality checks the authors perform, we look for incomplete annotations, missing explanations, and signs of misunderstanding judging from the provided reasoning.
If we find such a lack of quality, the respective annotator will be notified to re-do the annotation.
The quality checks are not used to bias annotators in their ratings but to ensure a complete and consistent dataset.
After the five weeks, we compute inter-annotator agreement scores (shown in \Cref{tab:krippendorffs_alpha_challenge,tab:krippendorffs_alpha_error}).
In case we observe a significant difference across annotators at this point, we have planned a dedicated meeting to discuss such cases with all annotators and a senior annotator to ensure that an understanding issue of the task or definition did not lead to the different ratings.

Annotators spend 43 minutes per sample, completing about seven samples daily.

\paragraph{Handling of scheduling conflicts:}
Given that our annotators have other commitments, we anticipate potential scheduling conflicts.
We allow flexibility for annotators to complete their samples beyond the week limit if needed, reserving a sixth week as a buffer. 
Despite these provisions, all annotators completed their assigned samples within the original weekly timeframes.

\section{Statistics on Annotation Labels}
\label{sec:appendix_annotation_labels}
In \Cref{tab:annotation_labels_statistics_challenges,tab:annotation_labels_statistics_errors}, we report statistics of the annotated dataset, showing how many are showing the respective challenge and error type.

\begin{table*}[]
\small
\centering
\begin{tabular}{l c c c c c c c c }
\toprule
 & \makecell{Spoken \\ Language} & \makecell{Speaker \\ Dynamic} & \makecell{Co- \\reference} & \makecell{Discourse \\ Structure} & \makecell{Contextual \\ Turn-Taking} & \makecell{Implicit \\ Context} & \makecell{Data \\ Scarcity} & \makecell{Low Inf. \\ Density}\\
\midrule
\# detections & \makecell{32\\(91\%)} & \makecell{35\\(100\%)} & \makecell{35\\(100\%)} & \makecell{34\\(97\%)} & \makecell{33\\(94\%)} & \makecell{5\\(14\%)} & \makecell{30\\(86\%)} & \makecell{35\\(100\%)}\\
\bottomrule
\end{tabular}
\caption{Statistics on the occurance of different challenges in the QMSum samples used as input. We report total number and corresponding percentage.}
\label{tab:annotation_labels_statistics_challenges}
\end{table*}

\begin{table*}[]
\small
\centering
\begin{tabular}{l c c c c c c c c }
\toprule
 & MI & Red & WR & IR & Hal & Inc & LI & SD\\
\midrule
LED & \makecell{31\\(89\%)} & \makecell{14\\(40\%)} & \makecell{4\\(11\%)} & \makecell{7\\(20\%)} & \makecell{9\\(26\%)} & \makecell{13\\(37\%)} & \makecell{3\\(9\%)} & \makecell{20\\(57\%)}\\

DialogLED & \makecell{33\\(94\%)} & \makecell{18\\(51\%)} & \makecell{4\\(11\%)} & \makecell{8\\(23\%)} & \makecell{8\\(23\%)} & \makecell{18\\(51\%)} & \makecell{3\\(9\%)} & \makecell{22\\(63\%)}\\

Pegasus-X & \makecell{34\\(97\%)} & \makecell{27\\(77\%)} & \makecell{14\\(40\%)} & \makecell{16\\(46\%)} & \makecell{13\\(37\%)} & \makecell{23\\(66\%)} & \makecell{10\\(29\%)} & \makecell{28\\(80\%)}\\

GPT3.5 & \makecell{34\\(97\%)} & \makecell{7\\(20\%)} & \makecell{3\\(9\%)} & \makecell{1\\(3\%)} & \makecell{5\\(14\%)} & \makecell{9\\(26\%)} & \makecell{9\\(26\%)} & \makecell{22\\(63\%)}\\

Zephyr & \makecell{34\\(97\%)} & \makecell{7\\(20\%)} & \makecell{3\\(9\%)} & \makecell{1\\(3\%)} & \makecell{8\\(23\%)} & \makecell{11\\(31\%)} & \makecell{11\\(31\%)} & \makecell{22\\(63\%)}\\

\bottomrule
\end{tabular}
\caption{Statistics on the occurance of different error types in the model-generated summaries. We report total number and corresponding percentage.}
\label{tab:annotation_labels_statistics_errors}
\end{table*}

\clearpage
\onecolumn
\hypertarget{annotation}{}
\citationtitle

\begin{bibtexannotation}
@inproceedings{kirstein-etal-2024-evaluation,
title        = {What's under the hood: Investigating Automatic Metrics on Meeting Summarization},
author       = {Kirstein, Frederic and Wahle, Jan Philip and Ruas, Terry and Gipp, Bela},
year         = {2024},
month        = {11},
booktitle    = {Findings of the Association for Computational Linguistics: EMNLP 2024},
publisher    = {Association for Computational Linguistics}
}\end{bibtexannotation}
\end{document}